\documentclass[conference]{IEEEtran}
\IEEEoverridecommandlockouts
\usepackage{cite}
\usepackage{amsmath,amssymb,amsfonts}
\usepackage{todonotes} 
\usepackage{algorithmic}
\usepackage{graphicx}
\usepackage{textcomp}
\usepackage{verbatim}
\usepackage{subcaption}
\usepackage[inkscapelatex=false]{svg}
\usepackage[colorlinks=true, urlcolor=blue]{hyperref}
\usepackage[font=footnotesize]{caption} 
\makeatletter

\renewcommand\subsubsection{\@secnumfont}{\bfseries}%
\renewcommand\subsubsection{\@startsection{subsubsection}{3}
  \z@{.5\linespacing\@plus.7\linespacing}{-.5em}%
  {\normalfont\bfseries}}
  
  \makeatother

\usepackage[numbers,sort&compress]{natbib} 
\usepackage{xcolor}
\def\BibTeX{{\rm B\kern-.05em{\sc i\kern-.025em b}\kern-.08em
    T\kern-.1667em\lower.7ex\hbox{E}\kern-.125emX}}

\makeatletter
\providecommand{\linespacing}{\baselineskip}
\makeatother

\begin{document}

\title{Spectral and Spatial Graph Learning for Multispectral Solar Image Compression {\footnotesize \textsuperscript{}}
}

\author{
Prasiddha Siwakoti\textsuperscript{\dag},
Atefeh Khoshkhahtinat\textsuperscript{\dag}
Piyush M. Mehta\textsuperscript{\ddag},\\
Barbara J. Thompson\textsuperscript{\S},
Michael S. F. Kirk\textsuperscript{\S},
Daniel da Silva\textsuperscript{\S}\\[1ex]
\textsuperscript{\dag}Department of Computer Science and Electrical Engineering, West Virginia University, WV, USA\\
\textsuperscript{\ddag}Department of Mechanical \& Aerospace Engineering, West Virginia University, WV, USA\\
\textsuperscript{\S}NASA Goddard Space Flight Center, MD, USA\\[1ex]
\texttt{\{ps00068, ak00043\}@mix.wvu.edu, \{piyush.mehta\}@mail.wvu.edu}\\
\texttt{\{barbara.j.thompson, michael.s.kirk, daniel.e.dasilva\}@nasa.gov}
}

\maketitle
\begin{abstract}

High-fidelity compression of multispectral solar imagery remains challenging for space missions, where limited bandwidth must be balanced against preserving fine spectral and spatial details. We present a learned image compression framework tailored to solar observations, leveraging two complementary modules: (1) the Inter-Spectral Windowed Graph Embedding (iSWGE), which explicitly models inter-band relationships by representing spectral channels as graph nodes with learned edge features; and (2) the Windowed Spatial Graph Attention and Convolutional Block Attention (WSGA-C), which combines sparse graph attention with convolutional attention to reduce spatial redundancy and emphasize fine-scale structures. Evaluations on the SDOML dataset across six extreme ultraviolet (EUV) channels show that our approach achieves a 20.15\% reduction in Mean Spectral Information Divergence (MSID), up to 1.09\% PSNR improvement, and a 1.62\% log-transformed MS-SSIM gain over strong learned baselines, delivering sharper and spectrally faithful reconstructions at comparable bits-per-\textbf{}pixel rates. The code is publicly available at 
\url{https://github.com/agyat4/sgraph}.
\end{abstract}

\begin{IEEEkeywords}
multispectral learned image compression, graph neural network, feature embedding, solar imagery
\end{IEEEkeywords}

\section{Introduction}
\label{sec:intro}


Multispectral images capture diverse range of physical phenomena across various spectral bands, providing essential information for fields such as remote sensing, atmospheric science, and astrophysics. However, their scientific utility hinges on preserving their high-fidelity representation, which imposes severe demands on storage and transmission systems. For example, NASA’s Solar Dynamics Observatory (SDO) generates approximately 1.4 terabytes of full-disk solar imagery per day, capturing thirteen spectral bands every 12 seconds \cite{2012SoPh}. The high volume of data overwhelms the limited telemetry and onboard storage capacities of space missions. While lossless compression preserves data quality, it typically reduces size by only 2–4 times. Lossy methods offer higher compression ratios but often compromise the fidelity required for scientific analysis. This balance between image fidelity and compression efficiency, known as the rate–distortion trade-off, forms the central motivation for the learned compression approach proposed in this paper

\begin{figure}[t]
    \centering
    \includegraphics[width= 0.9\linewidth]{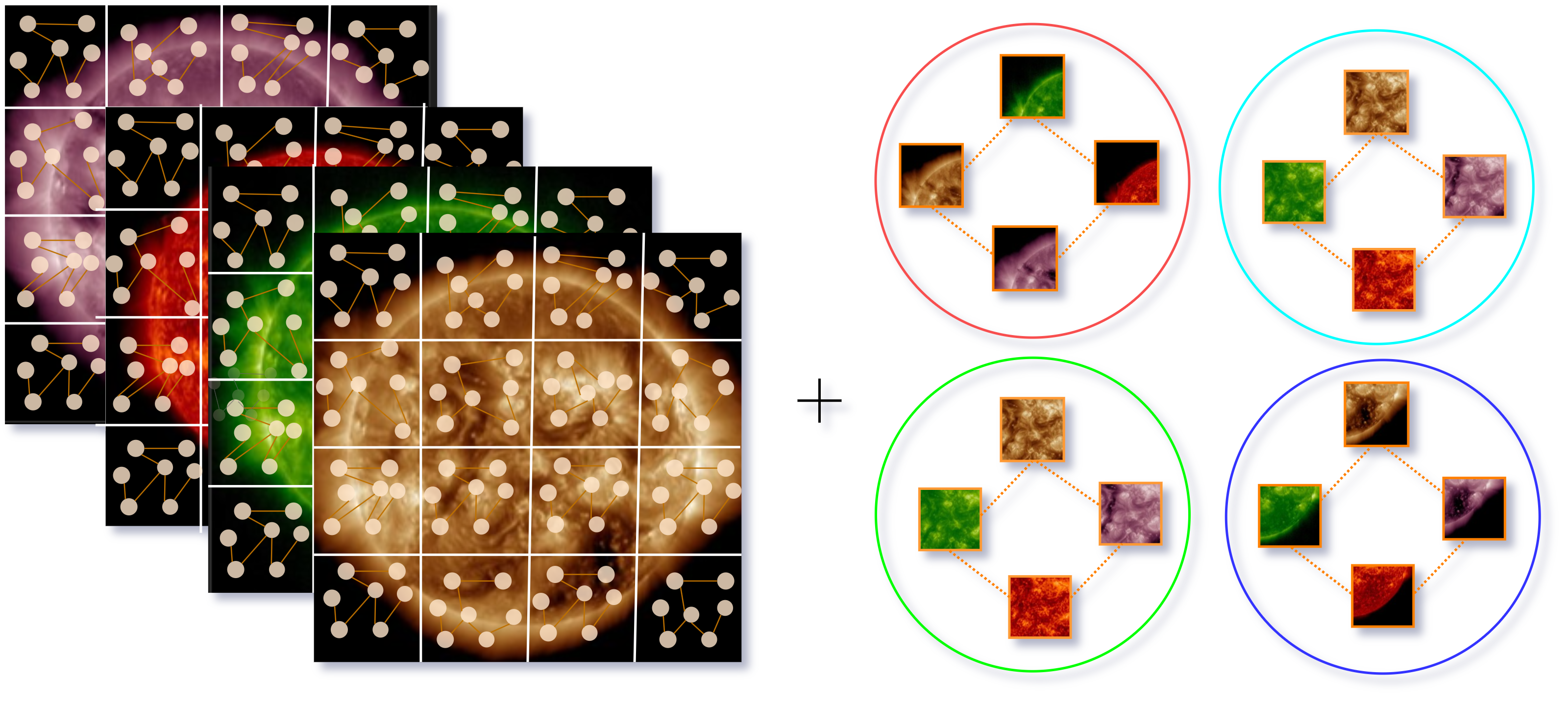}
    \caption{Multispectral solar images are partitioned into spatial windows. Graphs are constructed both within each window and across spectral channels to jointly capture spatial and spectral dependencies.}

    \label{fig:titlefig}
\end{figure}

Traditional image compression methods, such as JPEG \cite{125072} and JPEG2000 \cite{JPEG2000}, remain widely used but are not well-suited for multispectral or hyperspectral imagery, as they process each band independently and fail to exploit the cross-band correlations critical for preserving scientific fidelity. Learned image compression (LIC) methods overcome some of these limitations by jointly learning compact latent representations and entropy models for more efficient coding \cite{balle2016end}. Using convolutional networks \cite{balle2018variational}, \cite{minnen2018joint} and attention-based architectures \cite{cheng2020learned}, \cite{zou2022devil}, LIC has achieved strong results across many imaging domains and has even outperformed specialized codecs such as 3D-SPECK \cite{Tang2006} and 3D-SPIHT \cite{823937} on scientific datasets \cite{rs12213657}. However, most LIC approaches primarily target spatial redundancy, and even with advanced mechanisms such as windowed attention or recent graph-based models \cite{liu2023learned}, \cite{9710580}, the improvements in standard metrics such as PSNR and MS-SSIM have become increasingly marginal \cite{yang2023introduction}. To achieve further progress, particularly for bandwidth-constrained solar missions where fidelity across wavelengths is crucial, compression frameworks must explicitly capture the rich, non-uniform spectral correlations in multispectral data, building on while extending beyond 3D convolution and transformer-based approaches.

Multispectral images exhibit redundancy not only in spatial structure but also across spectral channels. While jointly modeling these dependencies is essential for effective compression, most learned methods address them only implicitly, which often fails to capture the structured correlations between specific bands. To address this, we introduce a graph-based multispectral compression framework that explicitly models inter-spectral relationships alongside spatial redundancies.

Our contributions are as follows:

\begin{itemize}
    \item We propose \textbf{Inter-Spectral Windowed Graph Embedding (iSWGE)}, a spectral module that represents bands as graph nodes and learns their relationships via edge embeddings. This targeted modeling preserves per-band semantics and leverages cross-band redundancy, resulting in enhanced spectral fidelity in solar imagery.

    \item We introduce a \textbf{Windowed Spatial Graph Attention (WSGA)} module that sparsifies local interactions to reduce redundancy and computational costs, paired with a Convolutional Block Attention Module (CBAM) that emphasizes informative spatial and channel-wise features. Working in concert with the spectral graph module iSWGE, \textbf{WSGA-C} enhances spatial representation efficiency and jointly contributes to superior spectral fidelity and overall compression performance.

    \item We validate our framework on the SDOML dataset, showing that the combined iSWGE and WSGA-C architecture consistently surpasses strong attention-based baselines, achieving superior spectral fidelity (20.15\% MSID reduction) and spatial quality gains (up to 1.09\% PSNR and 1.62\% MS-SSIM ) at identical bitrates, demonstrating bandwidth-efficient and scientifically reliable solar image compression. 
\end{itemize}

\section{Related Work}
\label{sec:format}

\subsection{Learning Image Compression}
\label{ssec:subhead}
Traditional image compression pipelines rely on transformation, quantization, and entropy coding. Modern learned image compression (LIC) frameworks adopt this structure but implement the components using deep neural networks, typically within a variational autoencoder (VAE) framework optimized for the rate–distortion trade-off. Reference \cite{balle2016end} introduced an end-to-end approach with nonlinear analysis and synthesis transforms and differentiable quantization , later extending it with a scale hyperprior to capture spatial variance patterns in the latent space, improving entropy modeling \cite{balle2018variational}. This framework was further advanced with autoregressive and hierarchical priors, enabling more accurate probability estimation and state-of-the-art compression rates \cite{minnen2018joint}.\\

\subsubsection{Transform Coding}
\label{sssec:subsubhead}

Convolutional neural networks (CNNs) remain the primary backbone for feature extraction in learned image compression (LIC), mapping images into compact latent representations. However, their limited receptive fields make capturing long-range dependencies challenging \cite{li2021involution}. Vision transformers, adapted from sequence modeling \cite{dosovitskiy2020image}, address this limitation but introduce substantial computational and memory costs. Swin Transformer architectures \cite{9710580} and \cite{zhu2022transformerbased} mitigate these costs by restricting self-attention to local windows, enabling non-local structure modeling with improved rate–distortion efficiency. Hybrid CNN–attention approaches \cite{cheng2020learned}, \cite{bai2022towards}, and \cite{Liu_2023_CVPR} further balance local and global modeling, enhancing feature stability while maintaining efficiency. Beyond pure attention mechanisms, Graph Neural Networks (GNNs) offer a flexible paradigm for modeling irregular structures in visual data \cite{zhou2020graph} and \cite{han2022visiongnnimageworth}. Recent attention-based graph designs \cite{spadaro2024gabic} integrate into hybrid LIC architectures \cite{zou2022devil}, replacing non-local attention with window-level graph operations. These treat patches as nodes and connect them via $k$-nearest neighbor (k-NN) graphs, selectively aggregating features to reduce redundancy while retaining essential relationships.\\

\subsubsection{Entropy Model}
\label{sssec:subsubhead}

Entropy modeling is a central component of learned image compression, as it estimates the probability distribution of quantized latents and directly determines coding efficiency. Early work employed a factorized Gaussian prior \cite{balle2016end}, enabling the end-to-end optimization of the rate–distortion trade-off. This was extended by introducing a scale hyperprior to capture spatial dependencies through hierarchical latent structures \cite{balle2018variational}. Autoregressive priors \cite{minnen2018joint} further improved compression by modeling sequential correlations, though they introduced substantial latency during encoding and decoding. To alleviate this, slice-based entropy models \cite{cheng2020learned} process latents channel-wise and reuse decoded slices as context, shortening causal contexts and improving parallelism, while relying less on explicit spatial context modeling.
ELIC \cite{he2022elic} builds on these advances by using uneven channel slicing combined with checkerboard spatial context, better capturing spatial and cross-channel correlations at the cost of greater computational complexity. More recent transformer-based designs, such as MLIC++ \cite{jiang2023mlic}, incorporate attention-driven mechanisms, including checkerboard attention and global feature aggregation, to efficiently model spatial and channel-wise priors. To further address long-range dependencies, Entroformer \cite{qian2022entroformer} employs top-$k$ self-attention with relative positional encoding to optimize probability estimation across the latent space.

\begin{figure*}[!t]
    \centering
    \includegraphics[width=\linewidth]{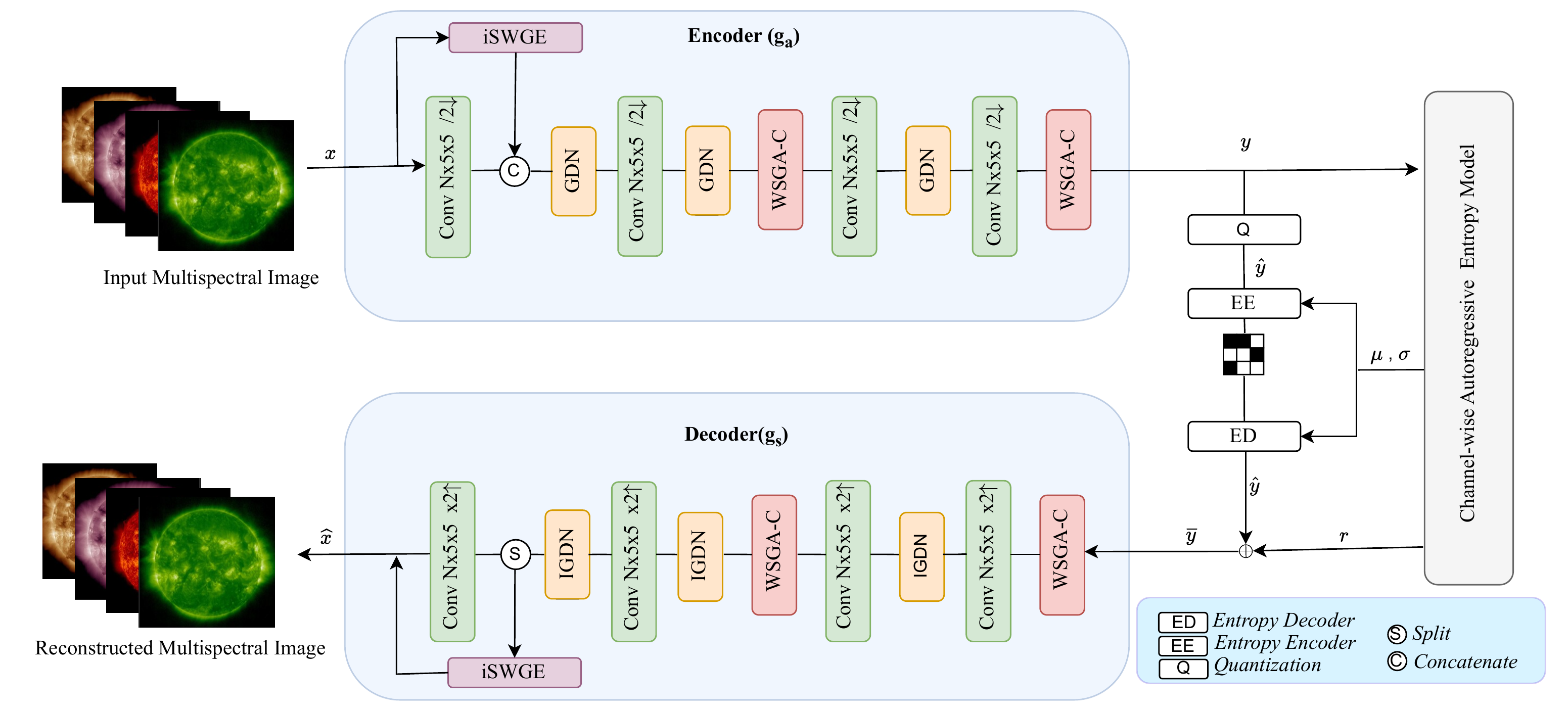}
    \caption{Overview of the proposed multispectral compression framework. The encoder $g_a$ fuses spectral features from iSWGE with spatial features via concatenation (C), followed by WSGA-C blocks to form compact latent representations. The latent $y$ is quantized (Q) and entropy-coded using a channel-wise autoregressive model. The decoder $g_s$ reconstructs spectral and spatial features through parallel paths: convolutional layers recover spatial layout, while iSWGE refines spectral embeddings. The two streams are averaged to produce the final output $\hat{x}$, ensuring fidelity in both spectral and spatial domains.}
    \label{fig:overall}
\end{figure*}

\subsection{Multispectral Image Compression}

Effective compression of multispectral imagery requires capturing both spatial and spectral correlations. Classical 3D wavelet-based codecs, such as 3D-SPECK and 3D-SPIHT \cite{Tang2006}, \cite{823937}, addressed this by jointly processing the spatial and spectral domains; however, recent learned approaches have surpassed them in rate–distortion performance. Recent Learned methods utilize multidirectional CNNs to fuse spectral and spatial features implicitly \cite{kong2022spectral} or employ attention mechanisms to adaptively weight spectral bands \cite{kong2022multi}. However, most of these approaches treat the spectral dimension as either a simple extension of the spatial domain or as a set of independent channels, overlooking their structured, non-uniform relationships. This limitation motivates the need for architectures that explicitly model inter-band dependencies to achieve more efficient and faithful compression.

\subsection{SDO Imagery: Learned Compression Advances}
\label{ssec:subhead}

Launched in 2010, NASA’s Solar Dynamics Observatory (SDO) monitors solar activity, including flares and coronal mass ejections (CMEs), with high temporal and spatial resolution \cite{Pesnell2012}. Its Atmospheric Imaging Assembly (AIA) captures 4096×4096 full-disk solar images across 10 ultraviolet wavelengths every 12 seconds, while the Helioseismic and Magnetic Imager (HMI) produces magnetograms and dopplergrams of the Sun’s magnetic field and interior \cite{2012SoPh}. Together, these instruments generate terabytes of high-cadence solar data daily. To support machine learning research on these data, the SDOML dataset \cite{galvez2019machine} curates AIA and HMI observations from 2010–2020, providing standardized data for tasks such as solar activity forecasting \cite{brown2022attention}, image restoration \cite{Zhang2023AttentionBasedDL}, and synthetic data generation \cite{dannehl2024experimental}.

Conventional compression approaches for SDO data remain limited. The Rice algorithm, being lossless, achieves only a 2-3 times compression factor \cite{Pesnell2012}, offering minimal bandwidth savings. JPEG2000 was tested for solar images \cite{fischer2017jpeg2000}, demonstrating limited storage and transfer gains, but requiring careful tuning of compression ratios to preserve the features of scientific interest, such as delicate coronal structures.
Recent work has shifted toward learned image compression for solar imagery. Reference \cite{zafari2024neural} introduced an adversarially trained network combining local and non-local attention with windowed convolutional attention blocks (CBAM), demonstrating significant rate–distortion improvements. A subsequent study \cite{zafari2023multi} proposed a transformer-based framework with token aggregation for multispectral compression, jointly encoding all wavelengths to leverage inter-band redundancies. However, this approach relies on the inherent feature aggregation of self-attention, without an explicit mechanism to model the structured, non-uniform relationships between spectral bands, which may limit its ability to fully exploit cross-band redundancies.

\section{Method}

\subsection{Compression Framework Overview}

We propose a learned image compression framework for multispectral solar images that jointly captures spatial and spectral redundancies through structured representation learning. Our model follows a transform coding paradigm comprising an analysis transform (encoder), a synthesis transform (decoder), and an entropy model for rate estimation.

In the encoder's early stage, a spectral graph module captures inter-spectral relationships by representing each band as a node and modeling edge embeddings between them. These spectral features mix with spatial features extracted in parallel via convolutional layers. To enhance spatial modeling, we incorporate two attention based approaches: a sparse windowed graph attention and a windowed CBAM. The windowed graph attention block sparsifies feature interactions by limiting aggregation to locally relevant nodes, mitigating redundancy introduced by fully connected attention. Complementing this, the windowed CBAM adaptively emphasizes spatially and spectrally informative regions, jointly promoting more efficient and informative representations for multispectral compression.

The latent representation is quantized and then modeled by a channel-wise autoregressive entropy model. The encoder transforms the input image $\mathbf{x} \in \mathbb{R}^{H \times W \times C}$ to a latent representation $\mathbf{y}$ and a hyper-latent $\mathbf{z}$. The quantized hyper-latent $\hat{\mathbf{z}}$ is then decoded into side information $\boldsymbol{\psi}$ by the hyper-synthesis transform $h_s$. Finally, the corrected latent tensor $\bar{\mathbf{y}}$ is used to reconstruct the output image $\hat{\mathbf{x}}$.
\begin{equation} \label{eq:transforms}
\begin{gathered}
    \mathbf{y} = g_a(\mathbf{x}; \phi) \\
    \mathbf{z} = h_a(\mathbf{y}) \\
    \boldsymbol{\psi} = h_s(\hat{\mathbf{z}}) \\
    \hat{\mathbf{x}} = g_s(\bar{\mathbf{y}}; \theta)
\end{gathered}
\end{equation}
The side information $\boldsymbol{\psi}$ contains base mean and scale tensors that provide the initial context for the entropy model. To facilitate this, the latent representation $\mathbf{y}$ divides into $s$ disjoint channel-wise slices. For each slice $\mathbf{y}_i$, a model predicts its Gaussian parameters $(\boldsymbol{\mu}_i, \boldsymbol{\sigma}_i)$ based on $\boldsymbol{\psi}$ and previously decoded slices $\hat{\mathbf{y}}_{<i}$. We then use a mean-subtracted approach to quantize the residual between the latent slice and its predicted mean. To improve reconstruction, an offset $\mathbf{r}_i$ is predicted and subtracted. The final probability model is a product of these Gaussian distributions, convolved with a uniform distribution to account for quantization noise.
\begin{equation} \label{eq:quant_entropy}
\begin{gathered}
    (\boldsymbol{\mu}_i, \boldsymbol{\sigma}_i) = \phi_i(\boldsymbol{\psi}, \hat{\mathbf{y}}_{<i}) \\
    \hat{\mathbf{y}}_i = \lfloor \mathbf{y}_i - \boldsymbol{\mu}_i \rceil + \boldsymbol{\mu}_i \\
    \bar{\mathbf{y}}_i = \hat{\mathbf{y}}_i - \mathbf{r}_i, \quad \text{where } \mathbf{r}_i = \rho_i(\hat{\mathbf{y}}_i, \boldsymbol{\mu}_i) \\
    p(\hat{\mathbf{y}} \mid \boldsymbol{\psi}) = \prod_{i} \mathcal{N}(\hat{\mathbf{y}}_i; \boldsymbol{\mu}_i, \boldsymbol{\sigma}_i^2) * \mathcal{U}(-0.5, 0.5)
\end{gathered}
\end{equation}
The proposed framework jointly optimizes bitrate and reconstruction quality under a rate-distortion objective. The full training objective minimizes the expected bitrate $R$ of both the latent and hyper-latent codes along with a distortion penalty $d(\cdot, \cdot)$, where the trade-off is controlled by $\lambda$.
\begin{equation}\label{eq:loss}
\begin{gathered}
 \mathcal{L} = \mathbb{E}_{\mathbf{x} \sim p(\mathbf{x})} \left[ -\log_2 p(\hat{\mathbf{y}} \mid \boldsymbol{\psi}) - \log_2 p(\hat{\mathbf{z}}) + \lambda \cdot d(\mathbf{x}, \hat{\mathbf{x}}) \right]
 \end{gathered}
 \end{equation}

\begin{figure}[!tp]
    \centering
    \begin{subfigure}[b]{0.48\columnwidth}
        \centering
        \includegraphics[width=\linewidth]{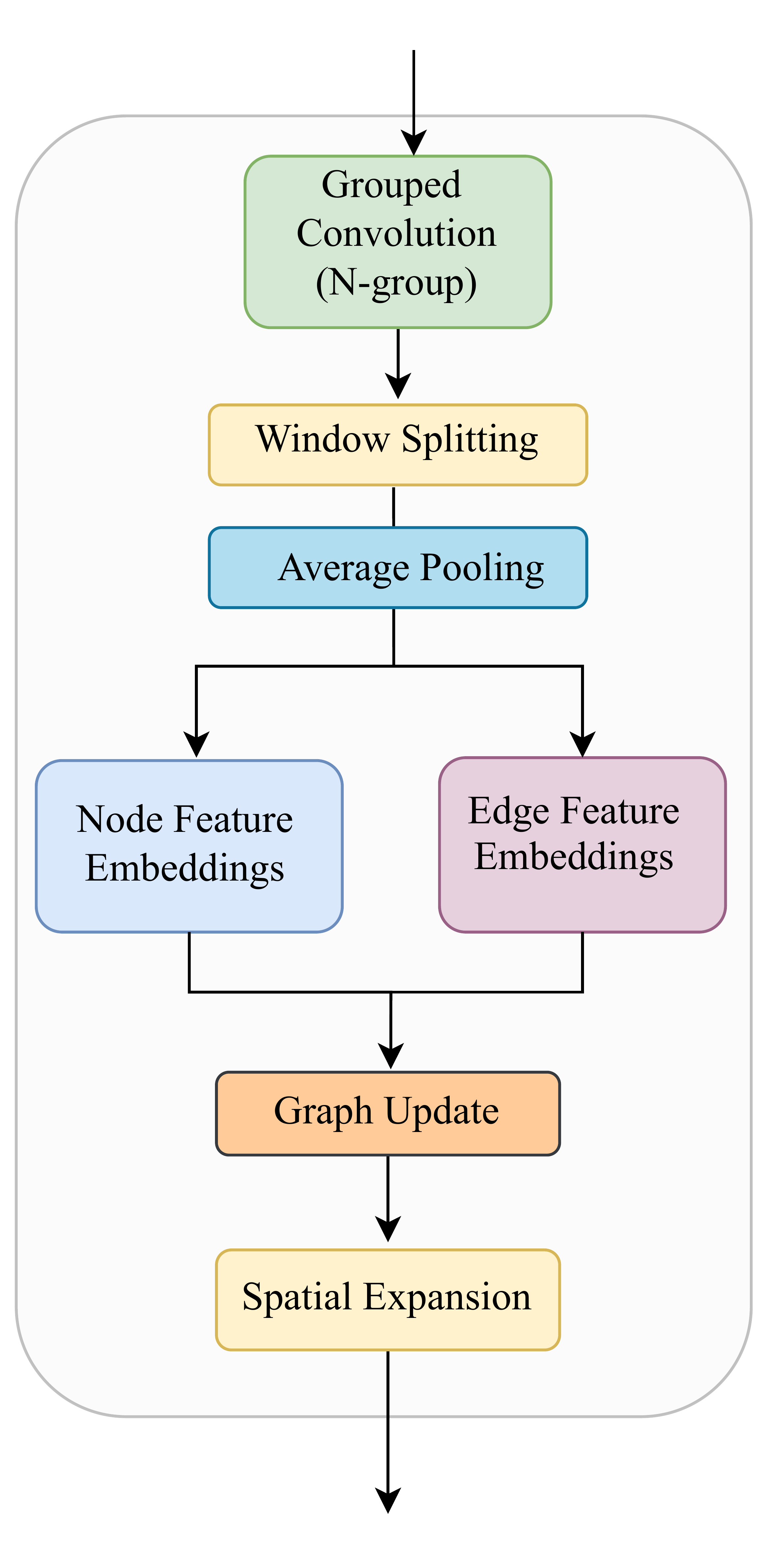}
        \caption{}
        \label{fig:spectral}
    \end{subfigure}
    \hfill
    \begin{subfigure}[b]{0.48\columnwidth}
        \centering
        \includegraphics[width=\linewidth]{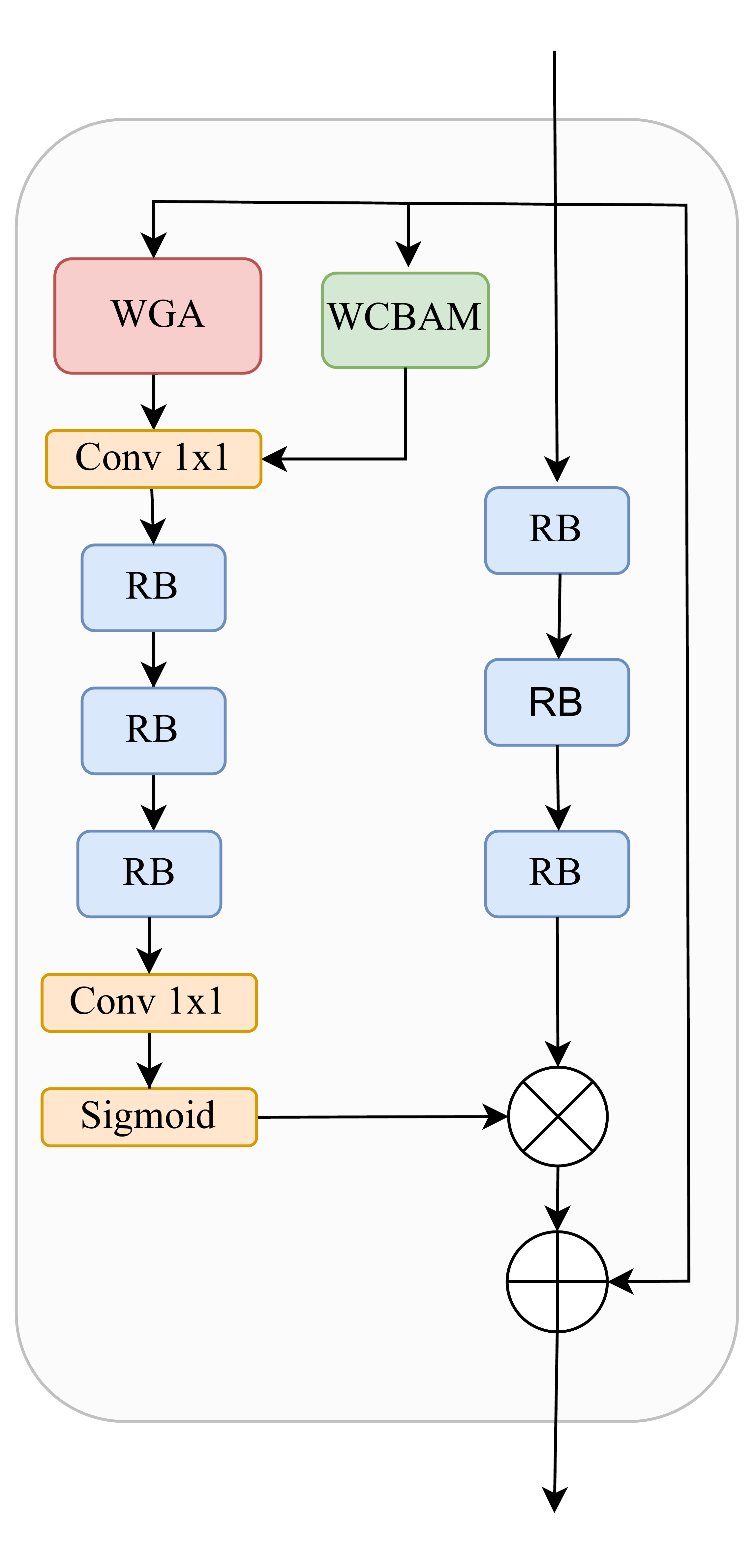}
        \caption{}
        \label{fig:spatial}
    \end{subfigure}
    \caption{Detail diagram of (a) the iSWGE module and (b) the WSGA-C module }
    \label{fig:grouped}
\end{figure}

\subsection{Inter-Spectral Windowed Graph Embedding (iSWGE)}

The \textbf{Inter-Spectral Windowed Graph Embedding (iSWGE)} module is designed to explicitly model the relational dependencies across spectral channels using a structured graph representation, leveraging correlations often missed by standard convolutional approaches. The process begins by applying a grouped convolution to the input image $\mathbf{x}$ with $g$  number of groups of each $f$ channels, which extracts a feature map $x' \in \mathbb{R}^{ H' \times W' \times (g\cdot f)}$ while preserving per-band semantics.
These feature maps are then partitioned into non-overlapping spatial windows. For each window, across the group, we construct a spectral graph. First, features in a window are summarized via average pooling into a descriptor matrix $\mathbf{X}^{(i)}\!\in\!\mathbb{R}^{g\times f}$. A lightweight MLP projects these into node embeddings $\mathbf{H}_v^{(i)}$, where each of the $g$ spectral groups becomes a node in the graph. We connect adjacent bands to form a cyclic graph, a static topology that is significantly more computationally efficient than constructing a dynamic k-NN graph and sparser than a fully-connected graph. To capture the local spectral contrast, a highly informative cue in solar imaging, edge features are formed by subtracting adjacent node descriptors and are projected into an embedding space by a shared MLP, creating edge embeddings $\mathbf{H}_e^{(i)}$.
\begin{gather}
    \mathbf{H}_v^{(i)}=\operatorname{MLP}\!\bigl(\mathbf{X}^{(i)}\bigr) \\
    \mathbf{H}_e^{(i)} = \operatorname{MLP}\!\bigl(|\mathbf{X}^{(i,j)}-\mathbf{X}^{(i,k)}|\bigr)
\end{gather}
The graph structure is defined by a binary incidence matrix $T^{(i)}\!\in\!\{0,1\}^{g \times |E^{(i)}|}$, which maps each edge to its two endpoint nodes, and by Laplacian-normalized adjacency matrices for the nodes ($\tilde{A}_v^{(i)}$) and edges ($\tilde{A}_e^{(i)}$) that describe their respective connectivity. To allow the spectral content (nodes) and their relationships (edges) to influence each other mutually, we refine these embeddings using a co-embedding mechanism adapted from CensNet~\cite{jiang2020co}. This iterative process creates richer, more context-aware representations. The update rule for a single layer $\ell$ is defined as:
\begin{equation}
\begin{aligned}
\mathbf{H}_v^{(\ell+1,i)} &=
\zeta\!\Bigl(\bigl[T^{(i)}\Phi(\mathbf{H}_e^{(\ell,i)}P_e)T^{(i)\!\top}\bigr]\!\odot\!\tilde{A}_v^{(i)}\,
\mathbf{H}_v^{(\ell,i)}W_v\Bigr) \\
\mathbf{H}_e^{(\ell+1,i)} &=
\zeta\!\Bigl(\bigl[T^{(i)\!\top}\Phi(\mathbf{H}_v^{(\ell+1,i)}P_v)T^{(i)}\bigr]\!\odot\!\tilde{A}_e^{(i)}\,
\mathbf{H}_e^{(\ell,i)}W_e\Bigr)
\end{aligned}
\end{equation}

where $\Phi(\cdot)$ is a diagonalization operator, $W_v, W_e$ are learnable weights, $P_v, P_e$ are learnable projection vectors, $\odot$ is the Hadamard product, and $\zeta$ is the ReLU function. Stacking three such layers yields deeply co-refined embeddings. Finally, the refined embeddings from each window are unshuffled back to their original spatial positions to reconstruct the full feature map. This spectrally aware feature map then concatenates with features from a parallel CNN branch, which captures complementary spatial patterns, thereby enriching the overall feature representation.

\subsection{Windowed Spatial Graph Attention and CBAM (WSGA-C)}

We introduce the \textbf{Windowed Spatial Graph Attention and CBAM (WSGA-C)} module to jointly address two significant challenges in LIC: (1) spatial redundant interactions in standard dense self-attention within local windows, and (2) ineffective selection of informative spatial and spectral features. WSGA-C addresses these issues by combining windowed sparse graph-based attention and windowed convolutional block attention in a parallel architecture, promoting efficient spatial representation learning and discriminative feature boosting.

To mitigate redundancy in local spatial attention, WSGA-C incorporates a spatial graph attention branch based on the formulation from GABIC~\cite{spadaro2024gabic}. Let the input feature tensor be $\tilde{\mathbf{x}} \in \mathbb{R}^{H'' \times W'' \times M}$. We divide this tensor into non-overlapping windows of size $P \times P$. Within each window, we treat each patch as a node in a local graph, $ G (V, E)$. The edge set $E$ is constructed dynamically in each iteration by computing the $k$-nearest neighbors ($\mathcal{N}(i)$) for each node based on feature similarity.
Each node is updated using a graph-based attention mechanism. Given a node $\tilde{\mathbf{x}}^{(i)} \in \mathbb{R}^{M}$, is updated  as:
\begin{equation}
\tilde{\mathbf{x}}_{\text{upd}}^{(i)} = \tilde{\mathbf{x}}^{(i)} + W_z \sum_{j \in \mathcal{N}(i)} \alpha_{i,j} W_g \tilde{\mathbf{x}}^{(j)},
\end{equation}
where $W_z$ and $W_g$ are learnable weight matrices. The attention coefficient $\alpha_{i,j}$ from node $i$ to its neighbor $j$ is computed via a softmax over the neighborhood, analogous to scaled dot-product attention:
\begin{equation}
\alpha_{i,j} = \frac{\exp \left( (W_\theta \tilde{\mathbf{x}}^{(i)})^\top W_\phi \tilde{\mathbf{x}}^{(j)} \right)}{\sum_{n \in \mathcal{N}(i)} \exp \left( (W_\theta \tilde{\mathbf{x}}^{(i)})^\top W_\phi \tilde{\mathbf{x}}^{(n)} \right)},
\end{equation}
where $W_\theta$ and $W_\phi$ are learnable projection matrices. When $k$ equals the total number of patches $N$, this formulation recovers standard local self-attention.

\begin{figure}[!tp]
    \centering
    \begin{subfigure}[b]{1.0\columnwidth}
        \centering
        \includegraphics[width=\linewidth]{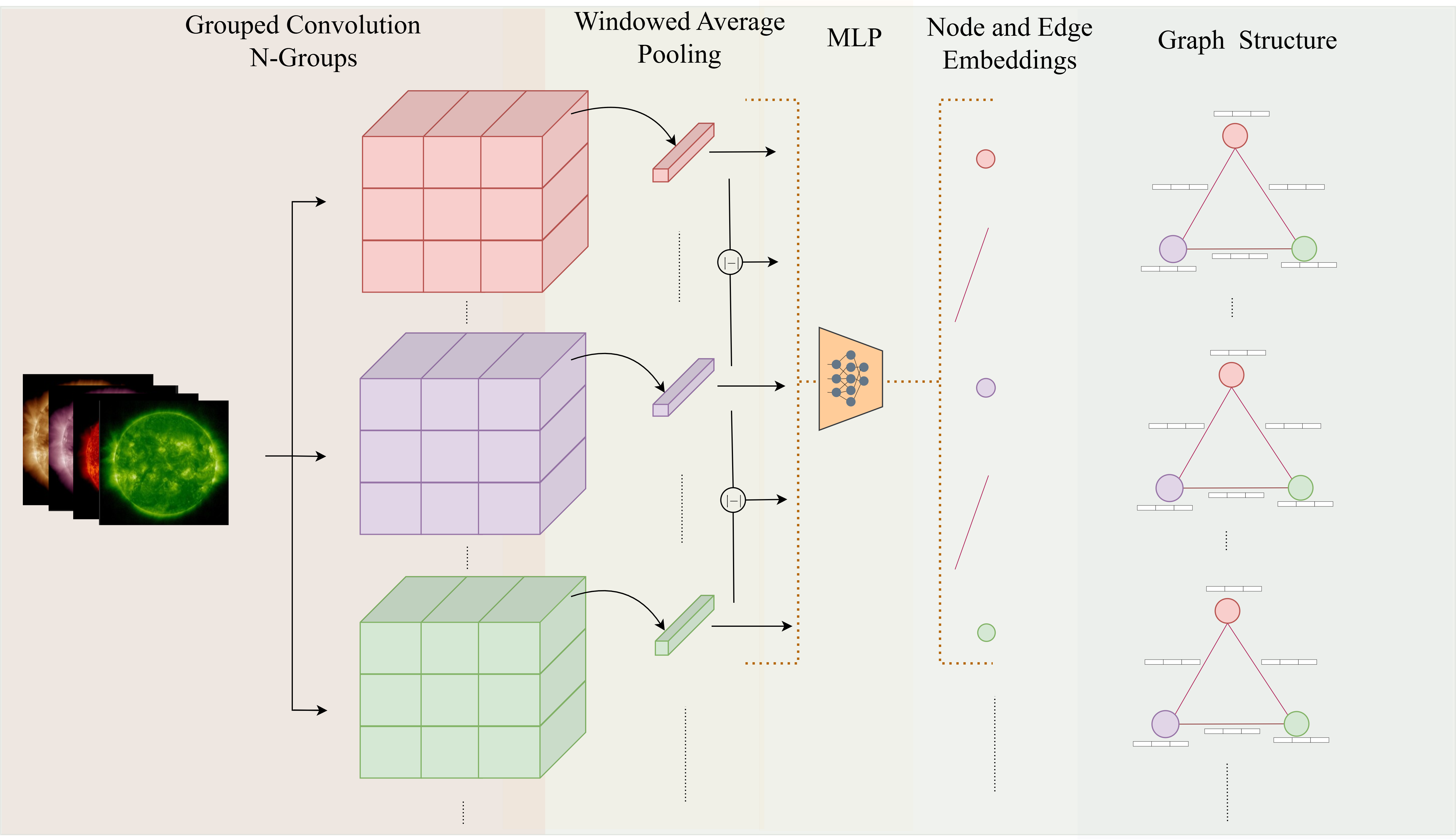}
        \label{fig:spectral}
    \end{subfigure}
    \caption{Spectral graph construction in iSWGE: grouped convolutions extract per-band features, pooled nodes form graph vertices, and edges encode differences for spectral graph processing.}
    \label{fig:spectralformation}
\end{figure}

To complement the spatial graph branch, WSGA-C applies a windowed CBAM that enhances features through channel and spatial attention. For a feature block $\tilde{\mathbf{x}}_{w} \in \mathbb{R}^{P \times P \times M}$, channel attention is first computed to obtain $A_{\mathrm{ch}}$, which reweights the feature channels to produce a channel-refined map $\tilde{\mathbf{x}}_{\mathrm{ch}}$. Spatial attention $A_{\mathrm{sp}}$ is then derived from $\tilde{\mathbf{x}}_{\mathrm{ch}}$ to emphasize informative regions, yielding the final output $\tilde{\mathbf{x}}_{\mathrm{wcbam}}$.The sequence of operations is:

\begin{equation}
\begin{gathered}
    A_{\mathrm{ch}} = \mathrm{sigmoid}\!\bigl(\mathrm{MLP}(\mathrm{Avg}(\tilde{\mathbf{x}}_{w})) + \mathrm{MLP}(\mathrm{Max}(\tilde{\mathbf{x}}_{w}))\bigr) \\
    \tilde{\mathbf{x}}_{\mathrm{ch}} = A_{\mathrm{ch}} \odot \tilde{\mathbf{x}}_{w} \\
    A_{\mathrm{sp}} = \mathrm{sigmoid}\!\bigl(\mathrm{Conv}([\mathrm{Avg}(\tilde{\mathbf{x}}_{\mathrm{ch}}), \mathrm{Max}(\tilde{\mathbf{x}}_{\mathrm{ch}})])\bigr) \\
    \tilde{\mathbf{x}}_{\mathrm{wcbam}} = A_{\mathrm{sp}} \odot \tilde{\mathbf{x}}_{\mathrm{ch}}
\end{gathered}
\end{equation}
Finally, instead of enforcing strict agreement between the two branches, we perform channel-wise concatenation of their outputs, followed by a learnable $1 \times 1$ convolution. This allows the network to adaptively integrate the relational interactions captured by the learned spatial graph with features emphasized by the CBAM.

\section{EXPERIMENTS}

\begin{figure*}[t]
    \centering
    \begin{subfigure}[b]{0.48\textwidth}
        \centering
        \includegraphics[width=\linewidth]{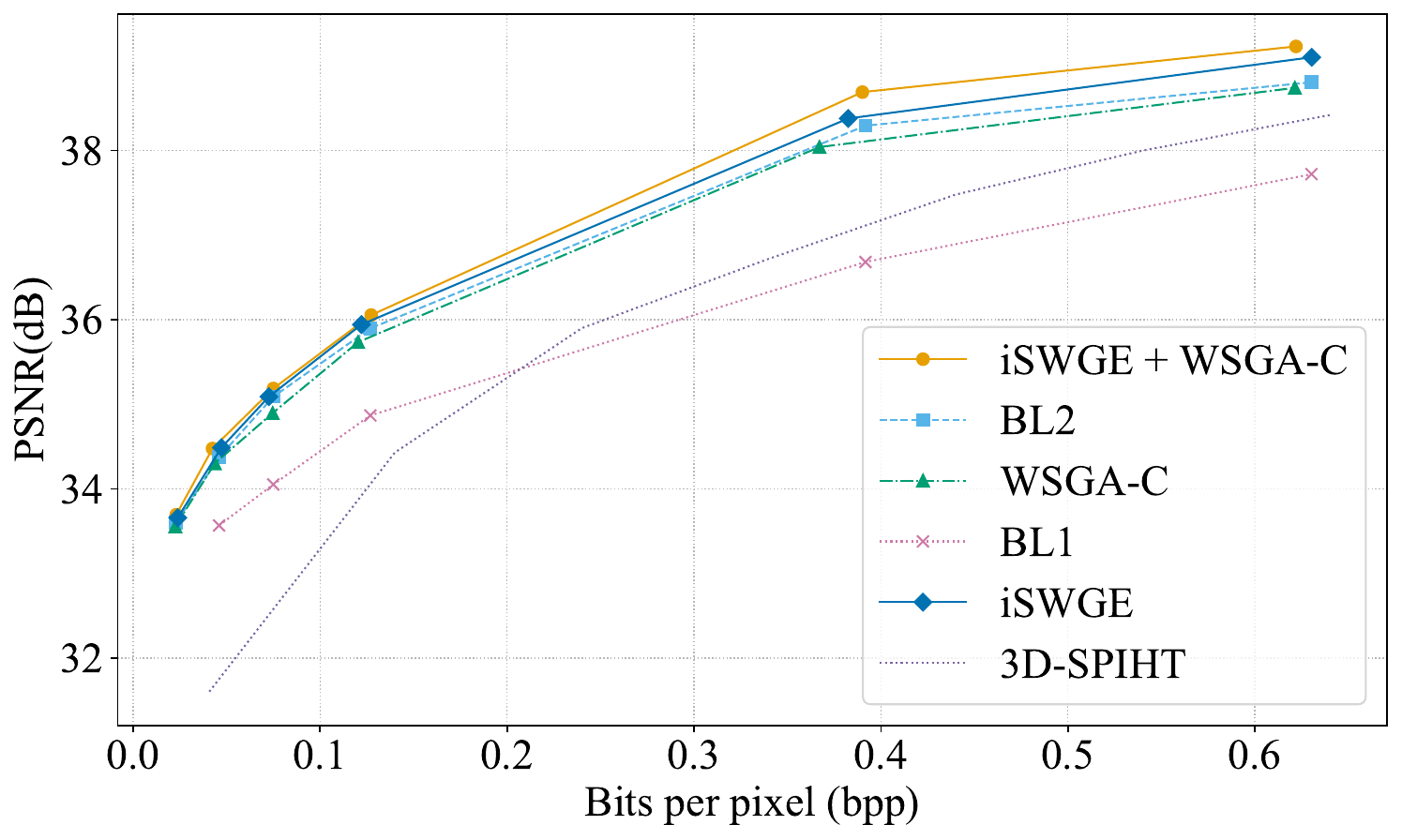}
        \caption{}
        \label{fig:psnr}
    \end{subfigure}
    \hfill
    \begin{subfigure}[b]{0.48\textwidth}
        \centering
        \includegraphics[width=\linewidth]{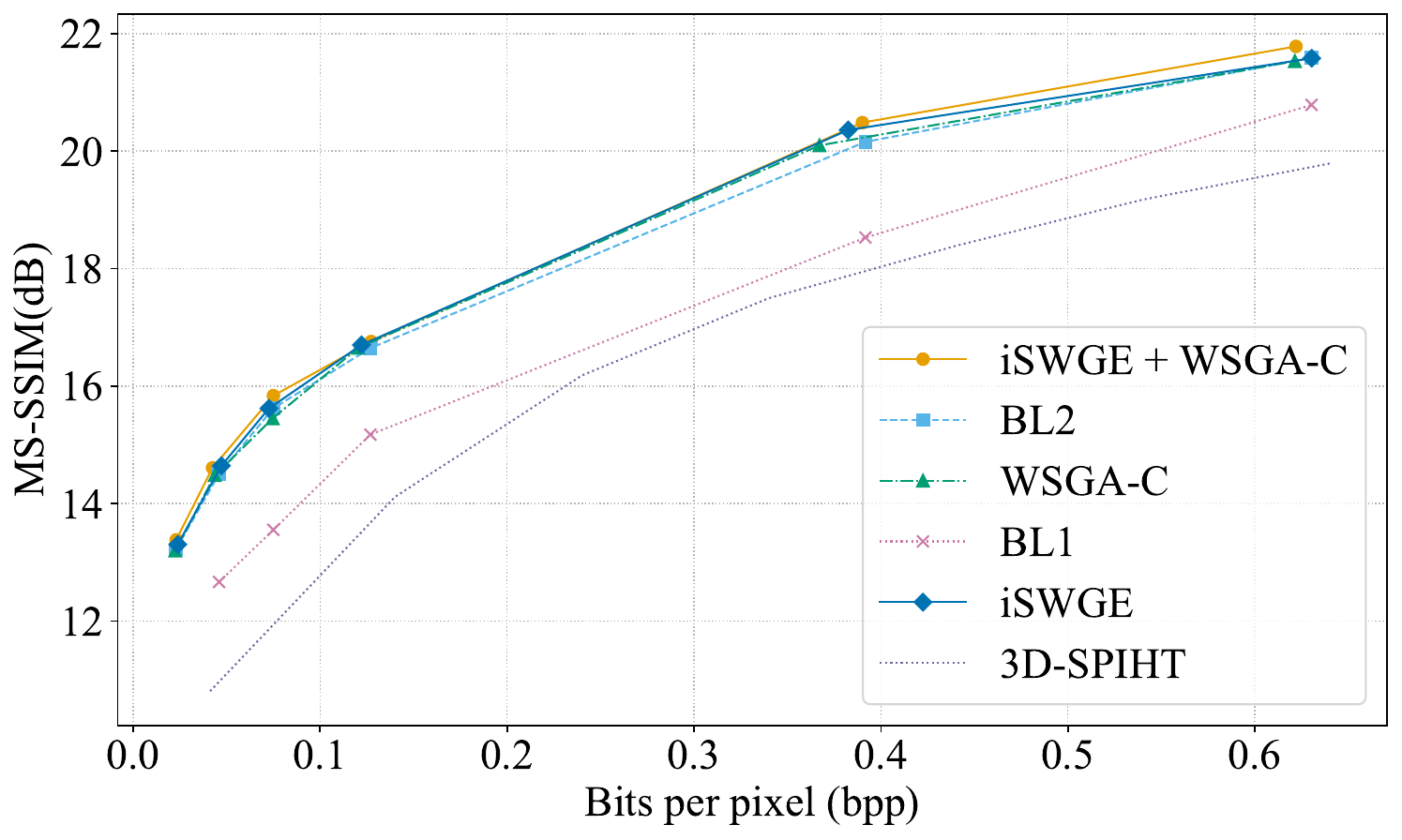}
        \caption{}
        \label{fig:msssim}
    \end{subfigure}
    \caption{Rate-distortion comparison of the proposed model against baselines: (a) PSNR vs. BPP and (b) log-transformed MS-SSIM vs. BPP. The (iSWGE + WSGA-C) model consistently outperforms BL1 and BL2 across bitrates, delivering higher spatial and perceptual quality.}
    \label{fig:rate-distortion}
\end{figure*}

\subsection{Experimental Setup}
\subsubsection{Dataset}

For our experiments, we use a curated subset of the SDOML dataset \cite{Pesnell2012}, focusing on six AIA EUV channels (94\,\AA, 131\,\AA, 171\,\AA, 193\,\AA, 211\,\AA, and 304\,\AA). These wavelengths capture emission features across a broad temperature range, from the transition region to active coronal structures. Preliminary tests also showed that including additional wavelength bands reduced cross-channel correlations and degraded compression performance. Following the protocol of \cite{zafari2023multi}, we resample all images to a uniform spatial resolution of $512 \times 512$ pixels. To reduce temporal correlation between samples and ensure diverse, independent training data, we reduce the temporal resolution by sampling one frame per hour from the native 6-minute cadence, which also improves computational efficiency during training. The dataset spans 2015–2018, and we partition it following \cite{salvatelli2019usingunetscreatehighfidelity}, we use the first eight months of each year for training and the remaining four months for testing. This temporal split prevents overfitting to specific phases of the solar cycle and evaluates the model’s generalization to unseen conditions. After filtering incomplete or corrupted entries, we compile a final dataset of \textbf{17,474 training} and \textbf{2,592 test} examples, each a six-channel stack of spatially aligned solar images.

\subsubsection{Training Setup}
We preprocess each input image by applying a logarithmic transformation to compress the dynamic range, followed by normalization and random cropping into \(256\times256\) patches. We train all models on an NVIDIA A30 GPU (24\,GB) using the Adam optimizer with a mini-batch size of 16 and an initial learning rate of \(10^{-4}\), adjusted dynamically via a Reduce-on-Plateau scheduler. Each model is trained for 250 epochs. To explore the rate–distortion trade-off, we train separate networks for each \(\lambda \in \{0.0018, 0.0048, 0.0085, 0.0150, 0.1000, 0.5000\}\). For WSGA-C, we set the K-NN value to 9 for graph construction, and for iSWGE, we limit the number of spatial windows to 16 to balance computational efficiency and feature aggregation.


\section{Evaluation}
\begin{figure}[htbp]
    \centering
    \includegraphics[width=\linewidth]{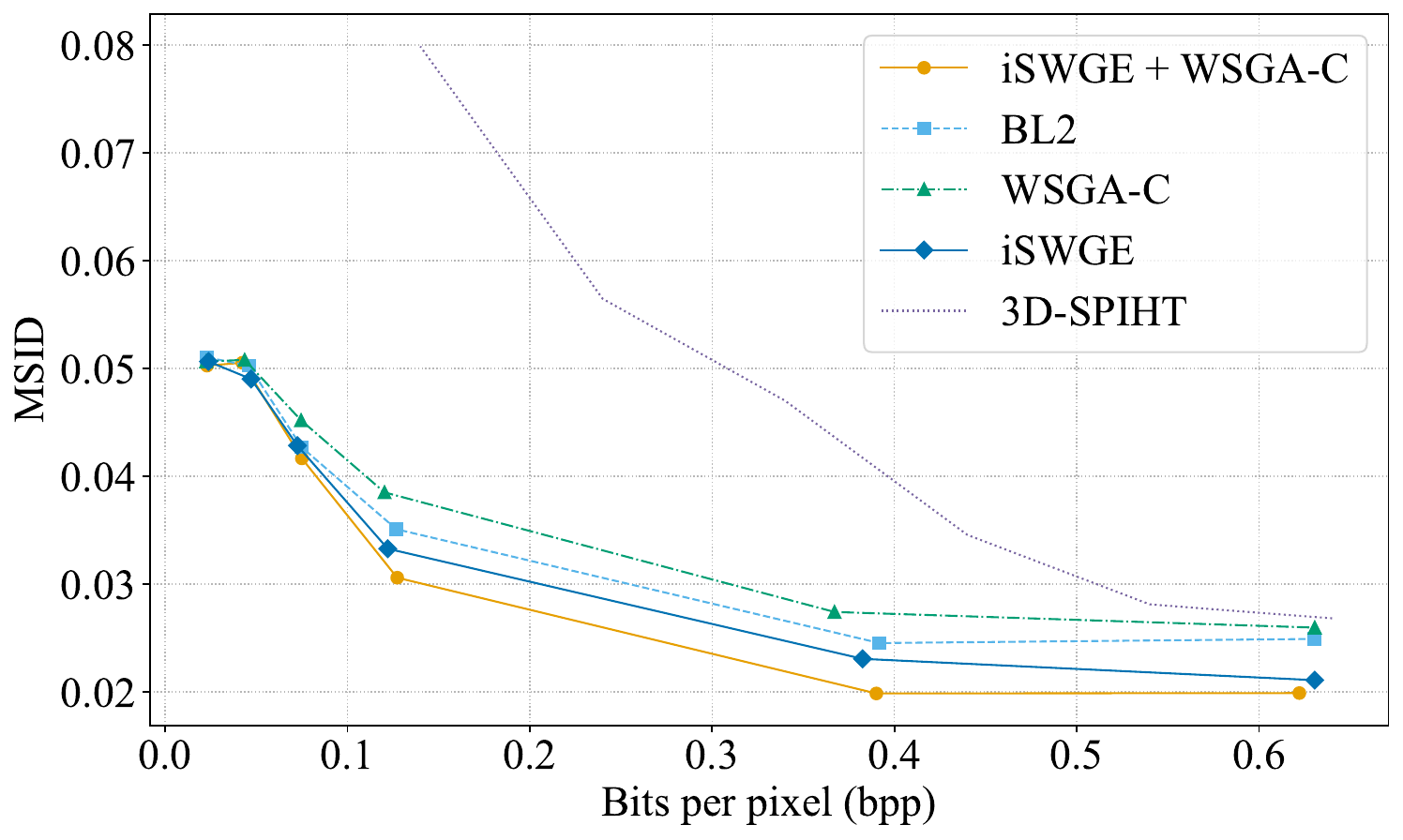}
    \caption{MSID vs. BPP, showing that (iSWGE + WSGA-C) achieves the lowest spectral divergence across bitrates. }
    \label{fig:msid}
\end{figure}

We evaluate compression performance using three metrics: Peak Signal-to-Noise Ratio (PSNR) for pixel-level fidelity, Multi-Scale Structural Similarity (MS-SSIM) \cite{1292216} for perceptual structure, and Mean Spectral Information Divergence (MSID) \cite{857802} for spectral accuracy, where lower MSID indicates better preservation of inter-band relationships. 

Baselines are adapted from the CNN-attention hybrid model in \cite{zafari2024neural}, evaluated without its GAN component, as adversarial training favors visual sharpness at the cost of PSNR and MS-SSIM. BL1 compresses each channel independently, providing a single-band reference, while BL2 jointly encodes all six channels by stacking them as a multi-channel input for the same encoder-decoder, where convolutional and attention layers implicitly capture cross-channel correlations. Even though the evaluation primarily focuses on learned baselines to assess architectural performance, the multispectral framework in \cite{zafari2023multi} is excluded, as it targets nine AIA channels and employs a distinct entropy coding scheme. All primary comparisons use compatible entropy models to isolate the impact of the transformation architecture. Additionally, we include the 3D-SPIHT codec to provide a representative traditional baseline for comparison.

\begin{table}[b]
\caption{Performance comparison of different model variants for encoding and decoding time, FLOPs and memory requirement.}
\centering
\resizebox{\columnwidth}{!}{%
\begin{tabular}{lcccc}
\hline
\textbf{Model Variant} & \textbf{Encode Time(ms)} & \textbf{Decode Time(ms)} & \textbf{FLOPs ($\times 10^9$)} & \textbf{Memory(MB)} \\ \hline
BL1                    & 77.48                    & 98.86                     & 349.28                        & 296.17               \\
BL2                    & 85.98                    & 110.30                    & 355.56                        & 303.32               \\
WSGA-C   & 98.20                    & 119.88                    & 354.69                        & 303.07               \\
iSWGE + WSGA-C  & 104.60                   & 131.52                    & 354.72                        & 304.11               \\ \hline
\end{tabular}%
}
\label{tab:performance}
\end{table}

\subsubsection{\textbf{Rate-Distortion Performance}}
Fig.~\ref{fig:rate-distortion} presents the rate-distortion performance of the proposed method compared to baselines BL1 and BL2 across a range of bitrates. Our spectral-spatial model consistently achieves higher PSNR values, with up to 0.4\,dB improvement over BL2 and nearly 1\,dB over BL1 at higher bitrates, highlighting its ability to preserve fine spatial details under compression better. In terms of perceptual quality, as shown in Fig.~\ref{fig:msssim}, the model delivers consistently higher MS-SSIM, with gains of up to 0.33\ dB in log-transformed MS-SSIM over BL2 and 1.06 dB over BL1, reflecting improved multi-scale structural coherence and visually superior reconstructions. For spectral fidelity, evaluated via MSID in Fig.~\ref{fig:msid}, our approach achieves a reduction of up to 20.15\% relative to BL2 at higher bitrates, demonstrating improved preservation of spectral signatures. MSID is only reported for jointly encoded models, as it measures divergence across normalized spectral distributions per pixel. For single-band methods like BL1, where channels are compressed and quantized independently with no shared bit allocation, MSID may not be meaningful.

\subsubsection{\textbf{Ablation Study}}
We conducted ablation studies to isolate the effects of the spectral graph module (iSWGE) and the spatial attention module (WSGA-C). As shown in Fig.~\ref{fig:psnr}, iSWGE alone improves the PSNR across all bitrates, validating its effectiveness in capturing structured inter-spectral correlations that are often overlooked by baseline convolutional encoders. In contrast, WSGA-C performs comparably to the baseline in terms of PSNR but does not exceed it. This is likely due to its sparsified graph-based attention, which prioritizes discriminative modeling over exhaustive spatial modeling. However, their combination (iSWGE + WSGA-C) yields the highest PSNR, confirming the effective synergy between spectral feature learning and localized spatial attention. Perceptual quality, as measured by MS-SSIM (Fig.~\ref{fig:msssim}), consistently improves with each module. This aligns with WSGA-C’s design objective of enhancing the salience of local features and inter-channel contrast through dual attention, resulting in smoother and more perceptually coherent reconstructions. 

An analysis of MSID (Fig.~\ref{fig:msid}) shows that iSWGE, when isolated, is most effective at reducing spectral divergence, underscoring its role in preserving bandwise semantic consistency. When combined with WSGA-C, the full model achieves the lowest MSID, indicating optimal spectral fidelity.

 As summarized in Table~\ref{tab:performance}, these improvements come with only a modest increase in computational demands and memory usage compared to BL2, as the sparse design of WSGA-C offsets some of the added complexity from iSWGE. This helps keep the full model practical while delivering superior spectral and spatial fidelity.

\section{CONCLUSION}
We developed a graph-based learned compression framework for multispectral solar imagery that overcomes the limitations of spatially focused approaches by explicitly capturing spectral dependencies and leveraging efficient spatial attention. The framework achieves lower spectral divergence and consistently improves spatial quality compared to strong learned baselines, enhancing overall rate-distortion performance while incurring only modest computational overhead. These findings demonstrate that the approach is a promising candidate for further evaluation by the solar physics and broader scientific imaging communities.


\section*{Acknowledgment}

 This research is based upon work supported by the National Aeronautics and Space Administration (NASA), via award number 80NSSC21M0322 under the title of \textit{Adaptive and  Scalable Data Compression for Deep Space Data Transfer Applications using Deep Learning } and the computational resources were provided by the WVU Research Computing Dolly Sods HPC cluster, which is funded in part by NSF OAC-2117575.

\bibliography{refs}

@ARTICLE{2012SoPh,
  author={J. R. Lemen and A. M. Title and D. J. Akin, {et al.}},
  journal={Solar Physics}, 
  title={The Atmospheric Imaging Assembly (AIA) on the Solar Dynamics Observatory (SDO)}, 
  year={2012},
  volume={275},
  number={1-2},
  pages={17--40},
  month={jan}
}

@ARTICLE{125072,
  author={G. K. Wallace},
  journal={IEEE Transactions on Consumer Electronics}, 
  title={The {JPEG} Still Picture Compression Standard}, 
  year={1992},
  volume={38},
  number={1},
  pages={xviii--xxxiv}
}

@book{JPEG2000,
  author={D. S. Taubman and M. W. Marcellin},
  title={{JPEG2000}: Image Compression Fundamentals, Standards and Practice},
  series={The Kluwer International Series in Engineering and Computer Science},
  publisher={Kluwer},
  year={2002}
}

@article{yang2023introduction,
  author={Y. Yang and S. Mandt and L. Theis and {et al.}},
  journal={Foundations and Trends in Computer Graphics and Vision}, 
  title={An Introduction to Neural Data Compression}, 
  year={2023},
  volume={15},
  number={2},
  pages={113--200}
}

@inbook{Tang2006,
  author={X. Tang and W. A. Pearlman},
  title={Three-Dimensional Wavelet-Based Compression of Hyperspectral Images}, 
  booktitle={Hyperspectral Data Compression}, 
  editor={G. Motta and F. Rizzo and J. A. Storer},
  pages={273--308},
  publisher={Springer US},
  address={Boston, MA},
  year={2006}
}

@article{823937,
  author={P. L. Dragotti and G. Poggi and A. R. P. Ragozini},
  journal={IEEE Transactions on Geoscience and Remote Sensing}, 
  title={Compression of Multispectral Images by Three-Dimensional {SPIHT} Algorithm}, 
  year={2000},
  volume={38},
  number={1},
  pages={416--428}
}

@article{rs12213657,
  author={C. Deng and Y. Cen and L. Zhang},
  journal={Remote Sensing}, 
  title={Learning-Based Hyperspectral Imagery Compression Through Generative Neural Networks}, 
  year={2020},
  volume={12},
  number={21},
  pages={3657}
}

@inproceedings{balle2016end,
  author={J. Ball{\'e} and V. Laparra and E. P. Simoncelli},
  title={End-to-End Optimization of Nonlinear Transform Codes for Perceptual Quality}, 
  booktitle={2016 Picture Coding Symposium (PCS)}, 
  pages={1--5},
  year={2016},
  organization={IEEE}
}

@article{balle2018variational,
  author={J. Ball{\'e} and D. Minnen and S. Singh and S. J. Hwang and N. Johnston},
  journal={arXiv preprint arXiv:1802.01436}, 
  title={Variational Image Compression with a Scale Hyperprior}, 
  year={2018}
}

@article{minnen2018joint,
  author={D. Minnen and J. Ball{\'e} and G. D. Toderici},
  journal={Advances in Neural Information Processing Systems}, 
  title={Joint Autoregressive and Hierarchical Priors for Learned Image Compression}, 
  year={2018},
  volume={31}
}

@inproceedings{cheng2020learned,
  author={Z. Cheng and H. Sun and M. Takeuchi and J. Katto},
  title={Learned image compression with discretized gaussian mixture likelihoods and attention modules},
  booktitle={Proceedings of the IEEE/CVF Conference on Computer Vision and Pattern Recognition},
  year={2020},
  pages={7939--7948}
}

@inproceedings{zou2022devil,
  author={R. Zou and C. Song and Z. Zhang},
  title={The Devil is in the Details: Window-Based Attention for Image Compression}, 
  booktitle={Proceedings of the {IEEE/CVF} Conference on Computer Vision and Pattern Recognition}, 
  pages={17492--17501},
  year={2022}
}

@inproceedings{liu2023learned,
  author={J. Liu and H. Sun and J. Katto},
  title={Learned Image Compression with Mixed Transformer-{CNN} Architectures}, 
  booktitle={Proceedings of the {IEEE/CVF} Conference on Computer Vision and Pattern Recognition}, 
  pages={14388--14397},
  year={2023}
}

@inproceedings{9710580,
  author={Z. Liu and Y. Lin and Y. Cao and H. Hu and Y. Wei and Z. Zhang and S. Lin and B. Guo},
  title={Swin Transformer: Hierarchical Vision Transformer Using Shifted Windows}, 
  booktitle={2021 {IEEE/CVF} International Conference on Computer Vision ({ICCV})}, 
  year={2021},
  pages={9992--10002},
  doi={10.1109/ICCV48922.2021.00986}
}

@inproceedings{spadaro2024gabic,
  author={G. Spadaro and A. Presta and E. Tartaglione and J. H. Giraldo and M. Grangetto and A. Fiandrotti},
  title={{GABIC}: Graph-Based Attention Block for Image Compression}, 
  booktitle={2024 {IEEE} International Conference on Image Processing ({ICIP})}, 
  year={2024},
  pages={1802--1808},
  organization={IEEE},
}

@article{kong2022spectral,
author={F. Kong and K. Hu and Y. Li and D. Li and X. Liu and T. Durrani},
journal={IEEE J. Sel. Top. Appl. Earth Observ. Remote Sens.},
title={A Spectral-Spatial Feature Extraction Method With Polydirectional CNN for Multispectral Image Compression},
year={2022},
volume={15},
number={},
pages={1-1},
doi={10.1109/JSTARS.2022.3158281}
}

@inproceedings{zafari2023multi,
  author={A. Zafari and A. Khoshkhahtinat and P. M. Mehta and N. M. Nasrabadi and B. J. Thompson and M. S. F. Kirk and D. {Da Silva}},
  booktitle={2023 International Conference on Machine Learning and Applications (ICMLA)}, 
  title={Multi-Spectral Entropy Constrained Neural Compression of Solar Imagery}, 
  year={2023},
  pages={1181--1188},
  organization={IEEE}
}

@article{jiang2020co,
  author = {X. Jiang and R. Zhu and S. Li and P. Ji},
  title = {Co-embedding of Nodes and Edges with Graph Neural Networks},
  journal = {IEEE Transactions on Pattern Analysis and Machine Intelligence},
  year = {2020},
  publisher = {IEEE},
}

@inproceedings{bai2022towards,
  author={Y. Bai and X. Yang and X. Liu and J. Jiang and Y. Wang and X. Ji and W. Gao},
  title={Towards end-to-end image compression and analysis with transformers},
  booktitle={Proceedings of the AAAI Conference on Artificial Intelligence},
  year={2022},
  volume={36},
  number={1},
  pages={104--112}
}

@inproceedings{li2021involution,
  author={D. Li and J. Hu and C. Wang and X. Li and Q. She and L. Zhu and T. Zhang and Q. Chen},
  title={Involution: Inverting the inherence of convolution for visual recognition},
  booktitle={Proceedings of the IEEE/CVF Conference on Computer Vision and Pattern Recognition},
  year={2021},
  pages={12321--12330}
}

@article{dosovitskiy2020image,
  author={A. Dosovitskiy},
  title={An image is worth 16x16 words: Transformers for image recognition at scale},
  journal={arXiv preprint arXiv:2010.11929},
  year={2020}
}

@inproceedings{zhu2022transformerbased,
  author={Y. Zhu and Y. Yang and T. Cohen},
  title={Transformer-based transform coding},
  booktitle={International Conference on Learning Representations},
  year={2022},
  url={https://openreview.net/forum?id=IDwN6xjHnK8}
}

@inproceedings{Liu_2023_CVPR,
  author={J. Liu and H. Sun and J. Katto},
  title={Learned image compression with mixed transformer-CNN architectures},
  booktitle={Proceedings of the IEEE/CVF Conference on Computer Vision and Pattern Recognition (CVPR)},
  year={2023},
  pages={14388--14397}
}

@article{zhou2020graph,
  author = {J. Zhou and G. Cui and S. Hu and Z. Zhang and C. Yang and Z. Liu and L. Wang and C. Li and M. Sun},
  title = {Graph Neural Networks: A Review of Methods and Applications}, 
  journal = {AI Open}, 
  year = {2020},
  volume = {1},
  pages = {57--81},
  publisher = {Elsevier},
}

@article{kong2022multi,
author={F. Kong and T. Cao and Y. Li and D. Li and K. Hu},
title={Multi-scale spatial-spectral attention network for multispectral image compression based on variational autoencoder},
journal={Signal Process.},
year={2022},
volume={198},
}

@inproceedings{he2022elic,
  author={D. He and Z. Yang and W. Peng and R. Ma and H. Qin and Y. Wang},
  title={ELIC: Efficient Learned Image Compression with Unevenly Grouped Space-Channel Contextual Adaptive Coding},
  booktitle={Proceedings of the IEEE/CVF Conference on Computer Vision and Pattern Recognition},
  year={2022},
  pages={5718--5727},
  
}

@inproceedings{jiang2023mlic,
  author={W. Jiang and J. Yang and Y. Zhai and P. Ning and F. Gao and R. Wang},
  title={MLIC: Multi-Reference Entropy Model for Learned Image Compression},
  booktitle={Proceedings of the 31st ACM International Conference on Multimedia},
  pages={7618--7627},
  year={2023}
}

@article{qian2022entroformer,
  author={Y. Qian and M. Lin and X. Sun and Z. Tan and R. Jin},
  title={Entroformer: A Transformer-Based Entropy Model for Learned Image Compression},
  journal={arXiv preprint arXiv:2202.05492},
  year={2022}
}

@article{Pesnell2012,
  author = {W. D. Pesnell and B. J. Thompson and P. C. Chamberlin},
  title = {The Solar Dynamics Observatory ({SDO})},
  journal = {Solar Physics},
  year = {2012},
  volume = {275},
  number = {1-2},
  pages = {3--15},
  doi = {10.1007/s11207-011-9841-3},
}

@article{galvez2019machine,
  author = {R. Galvez and D. F. Fouhey and M. Jin and A. Szenicer and A. Muñoz-Jaramillo and M. C. M. Cheung and P. J. Wright and M. G. Bobra and Y. Liu and J. Mason and others},
  title = {A Machine-Learning Data Set Prepared From the {NASA} Solar Dynamics Observatory Mission},
  journal = {The Astrophysical Journal Supplement Series},
  year = {2019},
  volume = {242},
  number = {1},
  pages = {7},
  publisher = {IOP Publishing},
}

@article{brown2022attention,
  author={E. J. E. Brown and F. Svoboda and N. P. Meredith and N. Lane and R. B. Horne},
  title={Attention-Based Machine Vision Models and Techniques for Solar Wind Speed Forecasting Using Solar EUV Images},
  journal={Space Weather},
  year={2022},
  volume={20},
  number={3},
  pages={e2021SW002976},
  publisher={Wiley Online Library}
}

@article{Zhang2023AttentionBasedDL,
  author={X. Zhang and L. Xu and Z. Ren and X. Yu and J. Li},
  title={Attention-Based Deep Learning Model for Image Desaturation of SDO/AIA},
  journal={Research in Astronomy and Astrophysics},
  year={2023},
  volume={23},
  url={https://api.semanticscholar.org/CorpusID:258729193}
}

@article{dannehl2024experimental,
  author={M. Dannehl and V. Delouille and V. Barra},
  title={An Experimental Study on EUV-to-Magnetogram Image Translation Using Conditional Generative Adversarial Networks},
  journal={Earth and Space Science},
  year={2024},
  volume={11},
  number={4},
  pages={e2023EA002974},
  publisher={Wiley Online Library}
}

@article{fischer2017jpeg2000,
  author={C. E. Fischer and D. Müller and I. {De Moortel}},
  title={{JPEG2000} Image Compression on Solar {EUV} Images},
    journal={Solar Physics},
  year={2017},
  volume={292},
  number={1},
  pages={16},
  publisher={Springer}
}

@article{zafari2024neural,
  author={A. Zafari and A. Khoshkhahtinat and J. A. Grajeda and P. M. Mehta and N. M. Nasrabadi and L. E. Boucheron and B. J. Thompson and M. S. F. Kirk and D. E. {da Silva}},
  title={Neural-Based Compression Scheme for Solar Image Data}, 
  journal={IEEE Transactions on Aerospace and Electronic Systems}, 
  year={2024},
  volume={60},
  number={1},
  pages={918--933},
  doi={10.1109/TAES.2023.3332056}
}

@ARTICLE{salvatelli2019usingunetscreatehighfidelity,
  author={Salvatelli, V. and Bose, S. and Neuberg, B. and dos Santos, L. F. G. and Cheung, M. and Janvier, M. and Baydin, A. G. and Gal, Y. and Jin, M.},
  title={Using U-Nets to create high-fidelity virtual observations of the solar corona},
  journal={arXiv preprint arXiv:1911.04006},
  year={2019}
}

@INPROCEEDINGS{1292216,
  author={Wang, Z. and Simoncelli, E.P. and Bovik, A.C.},
  booktitle={The Thrity-Seventh Asilomar Conference on Signals, Systems \& Computers, 2003}, 
  title={Multiscale structural similarity for image quality assessment}, 
  year={2003},
  volume={2},
  number={},
  pages={1398-1402 Vol.2},
  doi={10.1109/ACSSC.2003.1292216}}

@ARTICLE{857802,
  author={Chang, C.-I.},
  journal={IEEE Transactions on Information Theory}, 
  title={An information-theoretic approach to spectral variability, similarity, and discrimination for hyperspectral image analysis}, 
  year={2000},
  volume={46},
  number={5},
  pages={1927-1932},
  doi={10.1109/18.857802}}

@misc{han2022visiongnnimageworth,
      title={Vision GNN: An Image is Worth Graph of Nodes}, 
      author={Han, K. and Wang, Y. and Guo, J. and Tang, Y. and Wu, E.},
      year={2022},
      eprint={2206.00272},
      archivePrefix={arXiv},
      primaryClass={cs.CV},
      url={https://arxiv.org/abs/2206.00272}, 
}

\end{document}